\documentclass{article}
\usepackage{spconf,amsmath,graphicx}

\usepackage{enumitem}
\setlist{nosep, leftmargin=14pt}

\usepackage{mwe} 

\usepackage{todonotes}
\usepackage{enumitem}
\usepackage{graphicx}
\usepackage{subcaption}
\usepackage{amsmath}
\usepackage{bm}
\usepackage{url}
\usepackage{hhline}
\usepackage[backend=biber,
            style=ieee,
            citestyle=numeric-comp,
            maxnames=5,
            url=false,
            isbn=false]{biblatex}
\title{Deep Learning-based Point Cloud Registration for Augmented Reality-guided Surgery}
\name{Maximilian Weber$^{1}$ \qquad Daniel Wild$^{1}$ \qquad Jens Kleesiek$^{2}$ \qquad Jan Egger$^{1,2}$ \qquad Christina Gsaxner$^{1,2}$}
\address{$^{1}$ Institute of Computer Graphics and Vision, Graz University of Technology \\
$^{2}$ Institute for Artificial Intelligence in Medicine, University Hospital Essen (AöR)}

\begin{document}
%
\maketitle
\begin{abstract}
Point cloud registration aligns 3D point clouds using spatial transformations. It is an important task in computer vision, with applications in areas such as augmented reality (AR) and medical imaging. This work explores the intersection of two research trends: the integration of AR into image-guided surgery and the use of deep learning for point cloud registration. The main objective is to evaluate the feasibility of applying deep learning-based point cloud registration methods for image-to-patient registration in augmented reality-guided surgery. We created a dataset of point clouds from medical imaging and corresponding point clouds captured with a popular AR device, the HoloLens 2. We evaluate three well-established deep learning models in registering these data pairs. While we find that some deep learning methods show promise, we show that a conventional registration pipeline still outperforms them on our challenging dataset.

\end{abstract}

\section{Introduction}
\label{sec:intro}

Point cloud registration is a well-known problem in computer vision. The objective of point cloud registration is to find a spatial transformation that aligns a source point cloud with a target point cloud, maximizing their overlap. Point cloud registration has many applications in autonomous driving, robotics, object detection, pose estimation, augmented reality (AR), and medical imaging. One particular application for registration is the alignment of medical 3D images to the patient for image-guided surgery (IGS), the so-called image-to-patient registration. Recent research has explored the use of AR as a solution to address the challenges in traditional IGS, leading to AR-guided surgery (AR-GS)~\cite{1_CliRegTec}. Consequently, the task of image-to-patient registration becomes a focal point in the context of AR-GS. Currently, image-to-patient registration methods face challenges in meeting the precision demands of medical settings, operating efficiently on mobile hardware like AR devices, ensuring user-friendly interfaces for medical professionals with diverse technical expertise, and prioritizing patient comfort.

In recent years, there has been a significant body of research exploring the application of deep learning for point cloud registration, demonstrating impressive results in general point cloud alignment~\cite{5_SurveyDLBPCR}. However, to the best of our knowledge, no existing literature has specifically focused on using these methods for a medical AR application. This study aims to merge these two prevailing research directions — AR-GS and deep learning for point cloud registration. Our goal is to assess the effectiveness of contemporary deep learning-based methods for registering point clouds in the context of image-to-patient registration within AR-GS. 

We test recent point cloud registration models with real-world medical image data and point clouds acquired through a commercial, commonly used AR device, specifically the Microsoft HoloLens 2 (HL2). We created a dataset consisting of computed tomography (CT) patient data and corresponding point cloud images acquired with the AR headset. This dataset is significantly more challenging than common benchmarks, as the point clouds come from different sources (\textit{cross-source}). We selected three promising deep learning-based point cloud registration methods, Feature-metric Registration (FMR)~\cite{Huang_2020_CVPR}, PointNetLK Revisited~\cite{Li_2021_CVPR} and Deep Global Registration (DGR)~\cite{choy2020deep}, evaluate them on our dataset and compare them to a traditional registration method. In summary, this study underscores the promise of deep learning-based point cloud registration for image-to-patient registration in AR-GS.

\section{Related Work}

\subsection{Registration for augmented reality-guided surgery}
The utilization of 3D AR in medicine allows for the accurate presentation of crucial information directly within a healthcare professional's line of sight, providing valuable insights. Achieving precise spatial alignment between medical imaging data and the patient in medical scenarios involves employing various techniques for image-to-patient registration, each with its own set of advantages and disadvantages \cite{29_IGS,GSAXNER2023102757}.

\textbf{Manual registration} requires the user to manually place the medical image over the patient within their view through the AR device~\cite{9_ManSeg1,12_ManSeg4}. It is a technique that is commonly used because of its flexibility, full control of the user, and its ease of implementation due to the lack of additional equipment or complex algorithms. However, manual segmentation is time-consuming, prone to human errors, as it entirely depends on the experience of the operator, and lacks the ability to adapt in case of patient movement.

\textbf{Marker-based registration} uses artificial markers (fiducial markers) as reference points instead of natural landmarks on the human body to create spatial relationships between virtual and physical spaces for the registration process. Markers can be either detected and tracked outside-in using an external camera~\cite{22_MarkSeg10,23_MarkSeg11}, or inside-out~\cite{19_MarkSeg7,20_MarkSeg8}. 
Established libraries and toolboxes, such as Vuforia are available for easy deployment of these methods. They are fast and the movement of the patient can be tracked using markers. However, placement of markers may require invasive procedures involving risks and discomfort to the patient. The accuracy highly depends on visibility, lighting conditions and size of the marker~\cite{HLMarkerAcc}.

\textbf{Point cloud registration} is a viable alternative for AR-GS. The source point cloud of the patient's target anatomy can be easily extracted from MRI or CT imaging routinely acquired in the clinical routine. The target point cloud representing the patient in the physical space can be captured using external depth sensors~\cite{24_PointSeg1}, or inside-out, in case the AR device has depth sensing capabilities~\cite{28_PointSeg3,gsaxner2021augmented}.

\subsection{Deep learning-based point cloud registration}
Traditional point cloud registration usually involves a global alignment based on feature matching, followed by a local refinement using a variant of Iterative Closest Point (ICP)~\cite{Holz2015Registration3-D}. However, these methods face issues such as susceptibility to local minima and long runtimes due to their iterative nature. Deep learning techniques have shown immense promise for tasks involving 3D point clouds, including registration. Recent reviews provide exhaustive lists of current methods~\cite{6_SurveyPCRHuang,zhang2020deep,2_SurveyDLPCGuo}. They either focus on distinct steps from the traditional registration pipeline, such as feature computation, feature matching, or refinement or provide an end-to-end solution. Still, some key obstacles in deep learning-based registration remain, such as too small datasets, the size and dimension of the single data samples, and irregularity, clutter, and unstructuredness of 3D point clouds~\cite{2_SurveyDLPCGuo}. Particularly cross-source datasets, with varying densities, noise patterns, imperfect overlap, and a lack of perfectly matching points, pose difficulties for these models~\cite{6_SurveyPCRHuang}.

\section{Methods}
The overall scientific question our work tackles is: Is there currently a deep learning-based point cloud registration method that exhibits an "out-of-the-box" characteristic for seamless integration with AR-GS? Ideally, the final architecture should be an end-to-end method that can handle a medical cross-source dataset with many outliers. 

\subsection{Dataset and Pre-Processing}
The first step involves obtaining the dataset to simulate a genuine operational setting. The objective is to register the medical scan of the patient onto the patient on the operating table via the AR device, the HL2. To streamline the simulation process and avoid dependence on actual patients, CT scans from real patients and corresponding 3D-printed heads were employed in this study, taken from a publicly available medical AR facial data collection~\cite{DSFacialmodel}. We selected ten patients from this collection for our evaluation. 

To obtain the source point clouds, the skin surfaces were delineated from the PET/CT scans using thresholding and manual refinement, and points were extracted using the Marching Cubes algorithm. The pre-processing of CT scans was carried out using 3D Slicer. Since the resulting point clouds are very dense, we sub-sampled 10.000 data points from each scan, to approximate the average density of the target point cloud. We furthermore cut the first and last 10\% of points in z-direction (top and the bottom of the head) and the last 35\% of points in y-directio (back of the head), to increase the similarity between the source and target point clouds. 

\begin{figure}[h]
  \centering
  \begin{subfigure}[b]{0.23\textwidth}
    \centering
    \includegraphics[width=\textwidth,trim={2cm 0cm 2cm 0cm},clip]{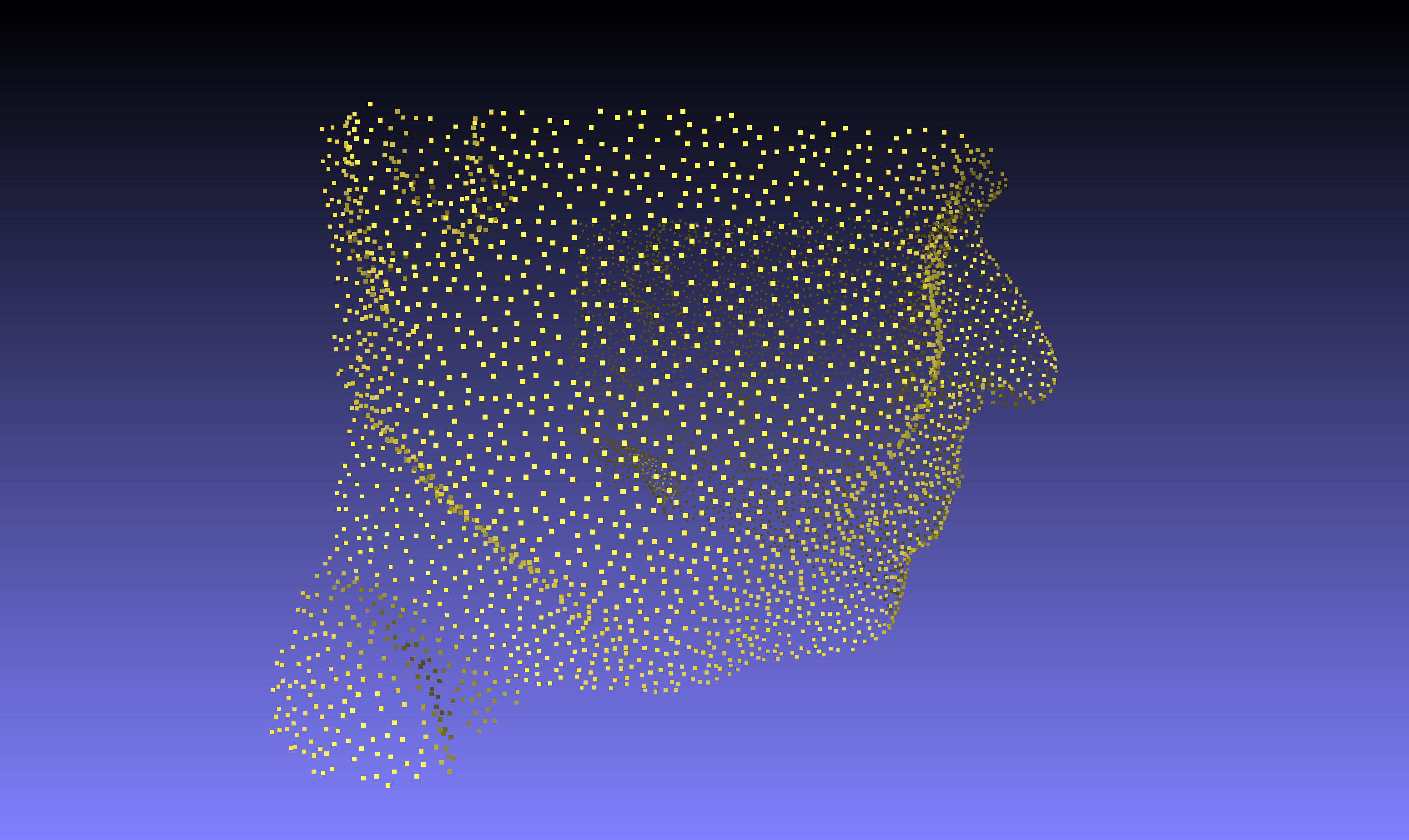}
    \caption{Source point cloud (CT)}
    \label{fig:source}
  \end{subfigure}
  \begin{subfigure}[b]{0.23\textwidth}
    \centering
    \includegraphics[width=\textwidth,trim={2cm 0cm 2cm 0cm},clip]{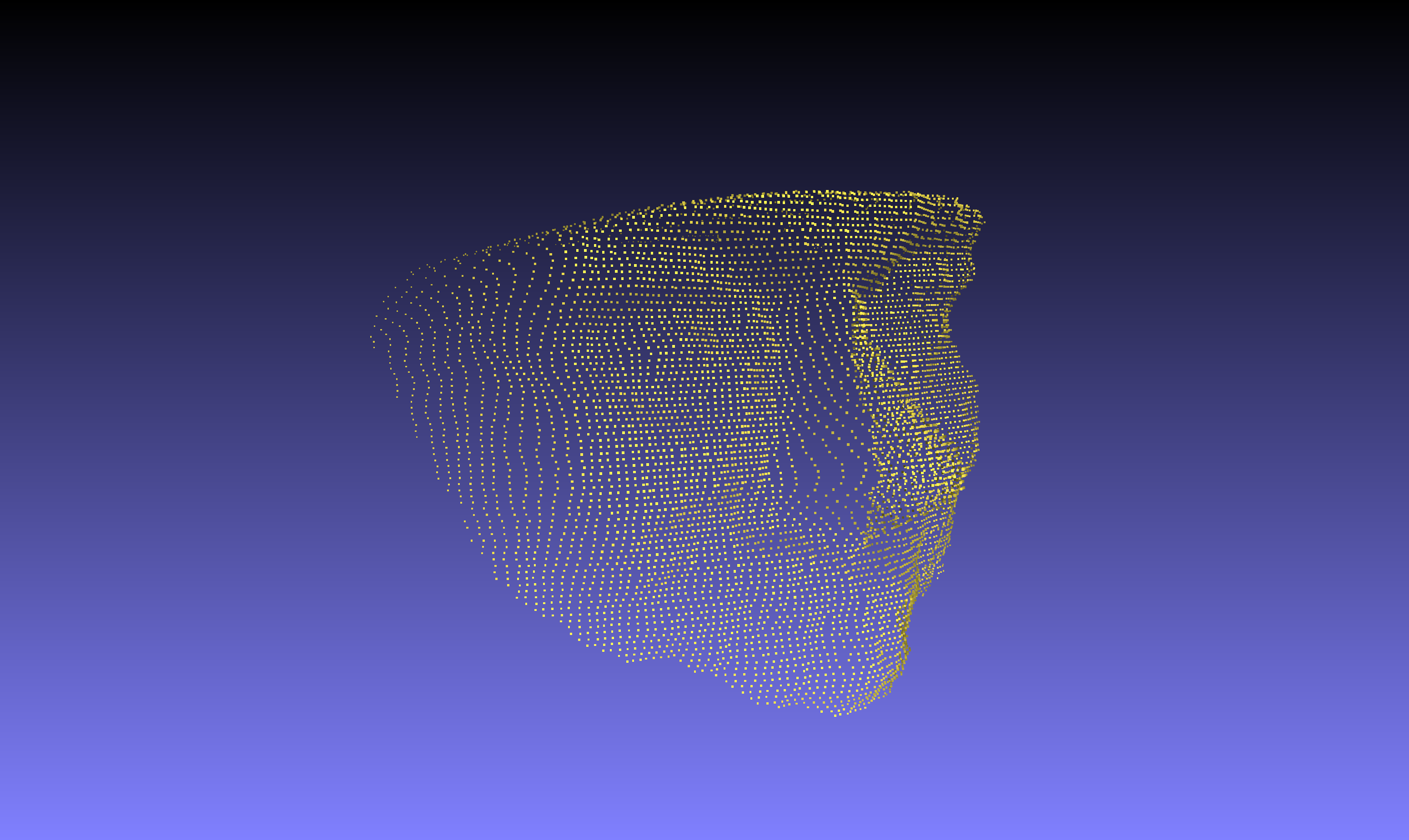}
    \caption{Target point cloud (HL2)}
    \label{fig:target}
  \end{subfigure}
  \caption{Point clouds after the pre-processing.}
  \label{fig:pre_processing_pics}
\end{figure}

To obtain the target data on which the source point clouds should be registered, we use the corresponding 3D-printed head phantoms. We recorded the patients' face phantoms from different perspectives with the depth sensor of the HL2. Specifically, we captured the "Depth AHaT" stream of the HL2, which delivers depth in the near range. We converted the depth images from the HoloLens to point clouds by re-projecting the depth pixels to 3D space using the camera parameters delivered by the device. To facilitate the registration process, we pre-process the target point cloud by removing unwanted points that might unduly perturb the registration. First, we clamp points within 25 cm from the camera, as artifacts frequently occur there from the reconstruction process. Then, we segment and remove the table on which the phantom is lying, by fitting a plane to the point cloud and removing all points in its proximity. Finally, we remove spurious single points. We use Python and the Open3D library for all point cloud pre-processing steps. Preprocessing of the target cloud runs fully automatically and in real-time ($\geq$ 30 frames per second). Both the source and target point clouds after pre-processing are depicted in Fig.~\ref{fig:pre_processing_pics}. We obtain ground truth between source and target using our benchmark method and manual refinement. In total, we collected 30 data pairs, with three perspectives of each patient.

\subsection{Benchmark method}
To obtain a benchmark to compare to the registrations obtained using deep learning-based methods, we employ a traditional registration pipeline that is known to work well~\cite{Holz2015Registration3-D}. We use a combination of the global registration and the ICP algorithm \cite{ICP}. ICP is a local registration method, which means it works very well when there is already a good alignment between the two point clouds. Since the point clouds in our dataset can be far apart, ICP alone would converge to a local minimum. Therefore, a global registration method is needed as initialization. Our global registration consists of fast point feature histograms (FPFH)~\cite{FPFH} which are matched using random sample consensus (RANSAC)~\cite{RANSAC}.

\subsection{Selection of deep learning-based registration methods}
In this work, we select three established methods for 3D point cloud registration from the multitude of available algorithms in the current literature. We base our selection on performance in benchmarks, reported capabilities for generalization, and cross-source adaptability. FMR~\cite{Huang_2020_CVPR} is a semi-supervised end-to-end approach that is very interesting for our dataset due to its reported cross-source registration performance. PointNetLK Revisited~\cite{Li_2021_CVPR} reportedly deals well with real-world and unseen data. Similar to traditional methods, it performs registration iteratively, combining PointNet and Lucas-Kanade (LK), an iterative optimization algorithm for image alignment. DGR~\cite{choy2020deep} combines three networks, for feature computation and matching matching, pose optimization and pose refinement, into an end-to-end registration solution, making it practical for replacing the traditional pipeline. All three selected methods are open-source and provide pre-trained models, which we use as the basis for our experiments.

\section{Experiments and Results}

\subsection{Benchmark Methods}

\begin{figure}[h]
  \centering
  \begin{subfigure}[b]{0.23\textwidth}
    \centering
    \includegraphics[width=\textwidth,trim={4cm 2cm 3.5cm 2cm},clip]{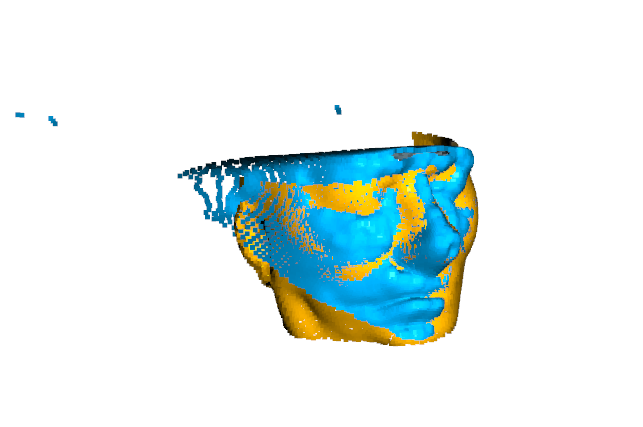}
    \caption{Global only registration.}
    \label{fig:global_alone_small}
  \end{subfigure}
  \begin{subfigure}[b]{0.23\textwidth}
    \centering
    \includegraphics[width=\textwidth,trim={2cm 1cm 2cm 1cm},clip]{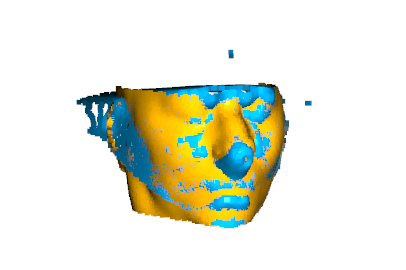}
    \caption{Global with ICP registration.}
    \label{fig:icp_global}
  \end{subfigure}
  \caption{Global registration alone and global registration with ICP initialization.}
  \label{fig:global_tests}
\end{figure}

First, we register each dataset using our benchmark method. A close examination of the global-only registration in Figure \ref{fig:global_alone_small} reveals that the registration is already reasonably accurate, though not precise enough. Combining global and ICP registration by utilizing the global registration as initialization for the ICP registration (global + ICP) yields the result shown in Figure \ref{fig:icp_global}. This outcome meets the required accuracy for our objectives and the available data.

\subsection{Deep Learning-based Methods}

\begin{figure}[ht]
  \centering
  \begin{subfigure}[b]{0.23\textwidth}
    \centering
    \includegraphics[width=\textwidth,trim={4cm 4cm 5cm 4cm},clip]{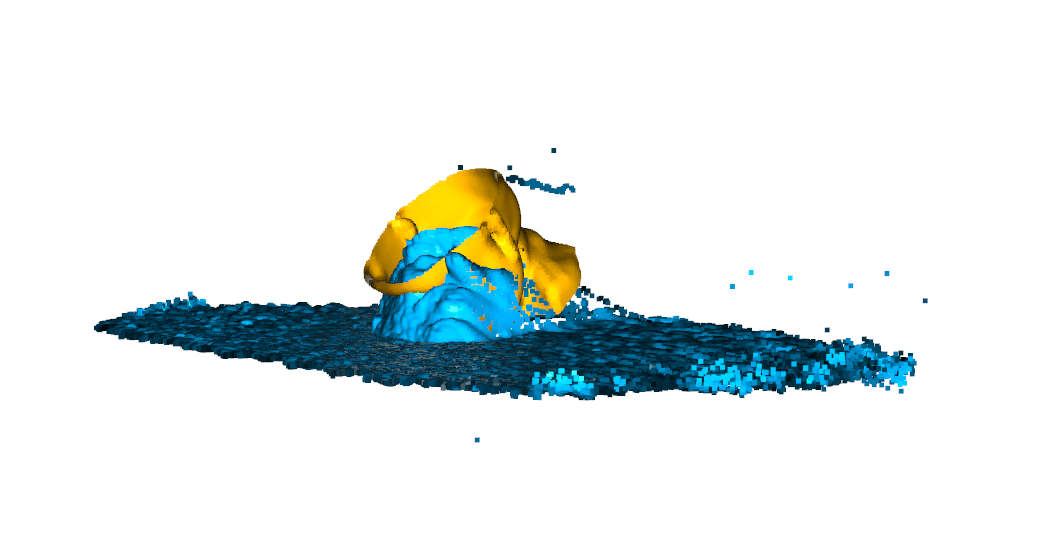}
    \caption{FMR}
    \label{fig:fmr_test}
  \end{subfigure}
  \begin{subfigure}[b]{0.23\textwidth}
    \centering
    \includegraphics[width=\textwidth,trim={4cm 4cm 5cm 4cm},clip]{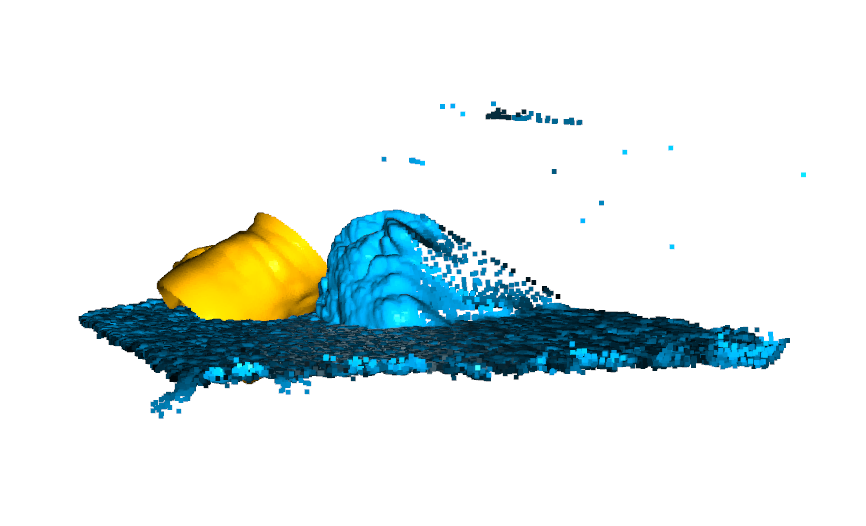}
    \caption{PointNetLK}
    \label{fig:pnlk_test}
  \end{subfigure}
  \caption{FMR~\cite{Huang_2020_CVPR} and PointNetLK Revisited~\cite{Li_2021_CVPR} registration result examples. Both methods fail to produce reasonable results for most of our cases.}
  \label{fig:fmr_pnlk_tests}
\end{figure}

Exemplary results with FMR and PointNetLK Revisited can be seen in Fig. \ref{fig:fmr_test} and Fig. \ref{fig:pnlk_test}, respectively. Evidently, these methods fail to find a satisfactory alignment for our dataset. To find out more about the possible reason of the bad performance of these methods, we conducted experiments registering the source point cloud to a rigidly transformed source point cloud, and the target point cloud to a rigidly transformed target point cloud, which yielded successful outcomes. We conclude that based on our findings, FMR and PointNetLK Revisited cannot cope with the inherent dissimilarity between the source and target point clouds in our dataset, stemming from the use of distinct sensors for capturing them.

\begin{figure}[ht]
  \centering
  \begin{subfigure}[b]{0.23\textwidth}
    \centering
    \includegraphics[width=\textwidth,trim={4cm 4cm 5cm 4cm},clip]{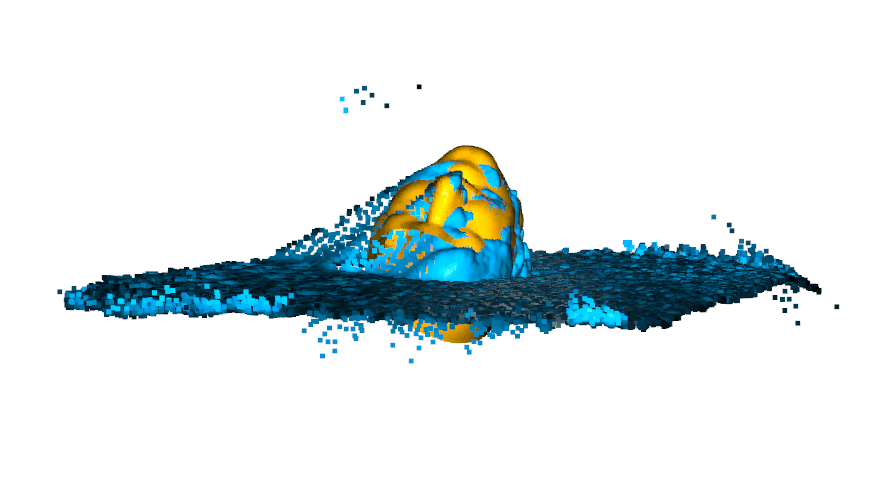}
    \caption{}
    \label{fig:dgr_owndata}
  \end{subfigure}
  \begin{subfigure}[b]{0.23\textwidth}
    \centering
    \includegraphics[width=\textwidth,trim={4cm 4cm 5cm 4cm},clip]{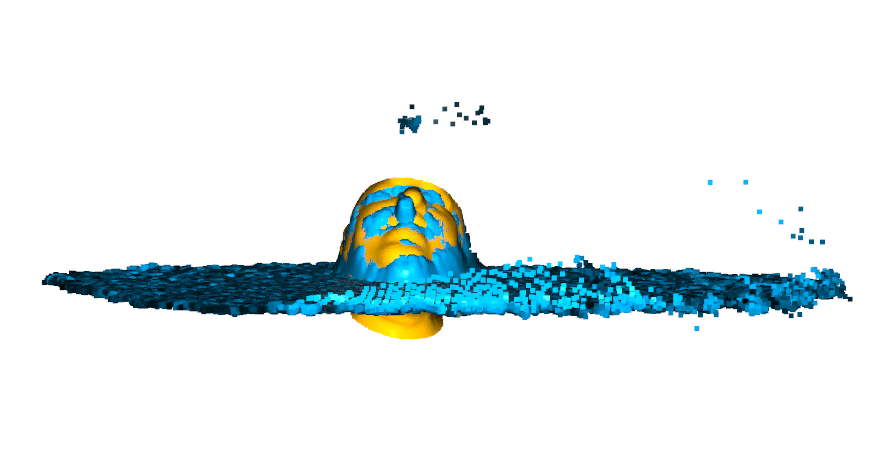}
    \caption{}
    \label{fig:dgr_owndata_2}
  \end{subfigure}
  \caption{DGR~\cite{choy2020deep} registration results. a) Presents a case with reasonable registration accuracy, b) shows a very promising result.}
  \label{fig:dgr_owndata_all}
\end{figure}

The results with DGR, as shown in Fig.~\ref{fig:dgr_owndata_all}, are encouraging, as in sample (a), up to very promising, as in sample (b). Hence, we fine-tuned this model on our own dataset. We split the data in training, validation and testing in ratios of 60\%, 30\% and 10\%. We set the batch size to 18 (i.e., the entire training set) and trained for 300 epochs using early stopping. The best model was identified after training for around 70 epochs. Finally, we compare the fine-tuned DGR (DGR FT) to the pre-trained DGR and our benchmark registration on the test set. We define the recall as the percentage of successful registrations. We consider a registration successful if the translation error is less than 0.4 cm and the rotation error is less than 15\textdegree. We furthermore report the error in translation (TE) and rotation (RE) in comparison to the ground truth registration for successful registrations. The results are summarized in Table~\ref{table:comparison}. Evidently, fine-tuning DGR significantly improves the recall, however, it has no positive effect on the errors. Still, the traditional benchmark of global + ICP registration shows the overall best performance. We do not compare against PointNetLK and FMR, as they had a recall of 0. 

One advantage of the DGR model is that results are reliably repeatable. If a good configuration has been found for a specific data sample, it will always successfully register this sample. For global registration or global + ICP registration, outcomes may differ for identical data samples because of the inherent randomness in the algorithms involved. While this demonstrates our DGR's stability, the recall and registration time also suggest that there is still room for improvement.  DGR, when trained with sufficient data, could be effective for medical applications. Comparatively, it was slower than global registration. However, compared to manual or semi-automatic registration commonly found in AR-GS, this delay wouldn't be problematic since it still presents a significant improvement. At the same time, DGR's accuracy is already acceptable for a lot of situations. The original DGR paper reports significantly shorter times (0.7 seconds), which could be attributed to hardware differences. We only tested on the CPU, as if the model is intended to run directly on the HL2 without performing the registration on an offsite server, hardware constraints must also be taken into consideration.

\begin{table}[htb]
\centering
\caption{Average registration results on our testing dataset. We test pre-trained DGR~\cite{choy2020highdimensional}, DGR fine-tuned on our dataset (DGR FT), feature-based global registration (Global)~\cite{Holz2015Registration3-D} and global registration refined with ICP (Gl + ICP)~\cite{ICP}.}
    \begin{tabular}{l|c|c|c|c}
    \textbf{Method} & \textbf{Recall} & \textbf{TE} (cm) & \textbf{RE} (°) & \textbf{T} (s) \\ 
    \hline
    DGR & 0.27 & 0.08 $\pm$ 0.97 & 4.71 $\pm$ 59.05 & 4.37  \\
    \hline 
    DGR FT & 0.41 & 0.09 $\pm$ 0.08 & 5.80 $\pm$ 5.59 & 4.40  \\ 
    \hline 
    Global & 0.30 & 0.18 $\pm$ 0.10 & 7.97 $\pm$ 3.08 & \textbf{1.55}  \\ 
    \hline 
    Gl + ICP & \textbf{0.60} & \textbf{0.05} $\pm$ 0.06 & \textbf{2.03} $\pm$ 2.08 & 1.63 \\
    \hline  
    \end{tabular}
\label{table:comparison}
\end{table}

\section{Conclusion}

We have successfully demonstrated that modern deep learning-based algorithms are capable of effectively capturing highly complex datasets. In our research, the data exhibit considerable variation in noise, density and distribution patterns. We observed that two techniques, FMR and PointNetLK, struggle with handling these disparities between our source and target point clouds. Despite preprocessing steps such as density alignment and noise reduction, the point clouds still retain distinct characteristics, incomplete alignment, and varied noise patterns owing to their different sources. Further research is required to validate these preliminary findings.

DGR, on the other hand, presents promising results. Based on our comprehensive evaluation, one way to enhance the accuracy of the method could be a hybrid approach that capitalizes on both DGR's adaptability to complex data and the high precision of ICP. The initialization could be done with the DGR algorithm and the refinement with ICP. This recommendation stems from our understanding of the strengths and limitations of each method and aligns with the broader goal of achieving accurate and efficient point cloud registration. 

Further research efforts could address the exploration of additional deep learning algorithms to register our own datasets and to fine-tune other deep learning methods to fit our specific data. For example, graph-based methods~\cite{fu2021robust} or other end-to-end algorithms~\cite{lu2019deepvcp} are promising. The potential benefits of using deep learning-based point cloud registration in IGS are significant, provided that ongoing efforts are directed toward refining these techniques to achieve optimal speed and accuracy consistent with the specific needs of medical applications. In addition, future studies could include the development of a dedicated application for HL2 that would allow direct implementation and testing of deep learning models. This would show how our research in the lab environment really performs in a real-world setting.

As the field of medical imaging continues to evolve, our research paves the way for the integration of advanced registration techniques into clinical applications. Potential use cases range from surgical planning to intraoperative guidance, where accurate and efficient registration of point cloud data can significantly impact procedural outcomes.

\section{Compliance with ethical standards}
\label{sec:ethics}
This research study was conducted retrospectively using
human subject data made available in open access by Egger et al. on figshare~\footnote{\url{https://doi.org/10.6084/m9.figshare.8857007.v2}}. Ethical approval was not required as confirmed by the license attached with the open access data.

\section{Acknowledgments}
\label{sec:acknowledgments}
This work received funding from the Austrian Science Fund (FWF) KLI 1044: \lq enFaced 2.0 - Instant AR Tool for Maxillofacial Surgery\rq, the REACT-EU project \lq KITE\rq (EFRE-0801977) and the Cancer Research Center Cologne Essen (CCCE). Authors declare no conflicts of interest.

\newpage
\printbibliography

@inproceedings{choy2020deep,
  title={Deep global registration},
  author={Choy, Christopher and Dong, Wei and Koltun, Vladlen},
  booktitle={CVPR},
  pages={2514--2523},
  year={2020}
}

@inproceedings{Li_2021_CVPR,
    author    = {Li, Xueqian and Pontes, Jhony Kaesemodel and Lucey, Simon},
    title     = {PointNetLK Revisited},
    booktitle = {CVPR},
    year      = {2021},
    pages     = {12763--12772}
}

@inproceedings{choy2020highdimensional,
    title={High-dimensional Convolutional Networks for Geometric Pattern Recognition}, 
    author={Christopher Choy and Junha Lee and Rene Ranftl and Jaesik Park and Vladlen Koltun},
    year={2020},
    booktitle={CVPR},
    pages={11224--11233}
}

@article{9_ManSeg1,
    author = {Gregory, Thomas and Gregory, Jules and Sledge, John and Allard, Romain and Mir, Olivier},
    title = {Surgery guided by mixed reality: presentation of a proof of concept},
    year = {2018},
    pages = {480--483},
    volume = {89},
    number= {5},
    journal = {Acta Orthopaedica}
}

@inproceedings{12_ManSeg4,
    author = {Nan Cui and Pradosh Kharel and Viktor Gruev},
    title = {Augmented reality with Microsoft HoloLens holograms for near infrared fluorescence based image guided surgery},
    booktitle = {SPIE BiOS},
    pages = {32--37},
    year = {2017}
}

@inproceedings{19_MarkSeg7,
    author = {Qian, Long and Zhang, Xiran and Deguet, Anton and Kazanzides, Peter},
    booktitle={MICCAI},
    year = {2019},
    pages = {74--82},
    title = {ARAMIS: Augmented Reality Assistance for Minimally Invasive Surgery Using a Head-Mounted Display}
}

@article{20_MarkSeg8,
    author = {Andreß, Sebastian and Johnson, Alex and Unberath, Mathias and Winkler, Alexander and Yu, Kevin and Fotouhi, Javad and Weidert, Simon and Osgood, Greg and Navab, Nassir},
    year = {2018},
    pages = {},
    title = {On-the-fly Augmented Reality for Orthopaedic Surgery Using a Multi-Modal Fiducial},
    volume={5},
    number={2},
    journal = {Journal of Medical Imaging}
}

@article{22_MarkSeg10,
    author = {Kuhlemann, Ivo and Kleemann, Markus and Jauer, Philipp and Schweikard, Achim and Ernst, Floris},
    year = {2017},
    pages = {184–-187},
    title = {Towards X-ray free endovascular interventions - Using HoloLens for on-line holographic visualization},
    volume = {4},
    number = {5},
    journal = {Healthcare Technology Letters}
}

@article{23_MarkSeg11,
    author = {Oliveira, Marcelo and Debarba, Henrique and Lädermann, Alexandre and Chagué, Sylvain and Charbonnier, Caecilia},
    year = {2019},
    title = {A Hand‐Eye Calibration Method for Augmented Reality Applied to Computer‐Assisted Orthopedic Surgery},
    volume = {15},
    number ={2},
    journal = {International Journal of Medical Robotics and Computer Assisted Surgery}
}

@article{24_PointSeg1,
    title={HoloLens-Based AR System with a Robust Point Set Registration Algorithm},
    author={Jong-Chih Chien and Yao-Ren Tsai and Chieh-Tsai Wu and Jiann-Der Lee},
    journal={Sensors (Basel)},
    year={2019},
    volume={19},
    number={16}
}

@inproceedings{28_PointSeg3,
    author="Gsaxner, Christina and Pepe, Antonio and Wallner, J{\"u}rgen and Schmalstieg, Dieter and Egger, Jan",
    pages = {236--244},
    title = {Markerless Image-to-Face Registration for Untethered Augmented Reality in Head and Neck Surgery},
    year = {2019},
    booktitle={MICCAI}
}

@article{29_IGS,
    author = {Grimson, W. and Kikinis, Ron and Jolesz, Ferenc and Black, Peter},
    year = {1999},
    pages = {62--69},
    title = {Image-Guided Surgery},
    volume = {280},
    number = {6},
    journal = {Scientific American}
}

@article{ICP,
  author={Besl, P.J. and McKay, Neil D.},
  journal={IEEE Transactions on Pattern Analysis and Machine Intelligence}, 
  title={A method for registration of 3-D shapes}, 
  year={1992},
  volume={14},
  number={2},
  pages={239-256}
}

@inproceedings{Huang_2020_CVPR,
    author = {Huang, Xiaoshui and Mei, Guofeng and Zhang, Jian},
    title = {Feature-Metric Registration: A Fast Semi-Supervised Approach for Robust Point Cloud Registration Without Correspondences},
    booktitle = {CVPR},
    year = {2020},
    pages = {11363--11371}
}

@article{RANSAC,
    author = {Fischler, Martin A. and Bolles, Robert C.},
    title = {Random Sample Consensus: A Paradigm for Model Fitting with Applications to Image Analysis and Automated Cartography},
    journal={Communications of the ACM},
    year = {1981},
    volume = {24},
    number = {6},
    pages = {381–-395}
}

@inproceedings{FPFH,
  author={Rusu, Radu Bogdan and Blodow, Nico and Beetz, Michael},
  booktitle={ICRA}, 
  title={Fast Point Feature Histograms (FPFH) for 3D registration}, 
  year={2009},
  pages={3212--3217}
}

@article{GSAXNER2023102757,
    title = {The HoloLens in medicine: A systematic review and taxonomy},
    journal = {Medical Image Analysis},
    volume = {85},
    year = {2023},
    author = {Christina Gsaxner and Jianning Li and Antonio Pepe and Yuan Jin and Jens Kleesiek and Dieter Schmalstieg and Jan Egger}
}

@article{HLMarkerAcc,
    author = {Pérez-Pachón, Laura and Sharma, Parivrudh and Brech, Helena and Gregory, Jennifer and Lowe, Terry and Poyade, Matthieu and Gröning, Flora},
    year = {2021},
    pages = {955-–966},
    title = {Effect of marker position and size on the registration accuracy of HoloLens in a non-clinical setting with implications for high-precision surgical tasks},
    volume = {16},
    number = {4},
    journal = {International Journal of Computer Assisted Radiology and Surgery}
}

@article{2_SurveyDLPCGuo,
  author={Guo, Yulan and Wang, Hanyun and Hu, Qingyong and Liu, Hao and Liu, Li and Bennamoun, Mohammed},
  journal={IEEE Transactions on Pattern Analysis and Machine Intelligence}, 
  title={Deep Learning for 3D Point Clouds: A Survey}, 
  year={2021},
  volume={43},
  number={12},
  pages={4338--4364}
}

@article{1_CliRegTec,
  author={Andrews, Christopher and Henry, Alexander B. and Soriano, Ignacio M. and Southworth, Michael K. and Silva, Jonathan R.},
  journal={IEEE Journal of Translational Engineering in Health and Medicine}, 
  title={Registration Techniques for Clinical Applications of Three-Dimensional Augmented Reality Devices}, 
  year={2021},
  volume={9},
  pages={1--14}
}

@article{5_SurveyDLBPCR,
    author = {Zhang, Zhiyuan and Dai, Yuchao and Sun, Jiadai},
    year = {2020},
    pages = {222--246},
    title = {Deep learning based point cloud registration: an overview},
    volume = {2},
    number = {3},
    journal = {Virtual Reality \& Intelligent Hardware}
}

@article{6_SurveyPCRHuang,
  author    = {Xiaoshui Huang and
               Guofeng Mei and
               Jian Zhang and
               Rana Abbas},
  title     = {A comprehensive survey on point cloud registration},
  journal   = {arXiv CoRR},
  volume    = {abs/2103.02690},
  year      = {2021},
  url       = {https://arxiv.org/abs/2103.02690}
}

@article{DSFacialmodel,
    author = {Gsaxner, Christina and Wallner, Jürgen and Chen, Xiaojun and Zemann, Wolfgang and Egger, Jan},
    year = {2019},
    title = {Facial model collection for medical augmented reality in oncologic cranio-maxillofacial surgery},
    volume = {6},
    journal = {Scientific Data}
}

@article{zhang2020deep,
  title={Deep learning based point cloud registration: an overview},
  author={Zhang, Zhiyuan and Dai, Yuchao and Sun, Jiadai},
  journal={Virtual Reality \& Intelligent Hardware},
  volume={2},
  number={3},
  pages={222--246},
  year={2020},
  publisher={Elsevier}
}

@article{Holz2015Registration3-D,
    title = {Registration with the Point Cloud Library: A Modular Framework for Aligning in 3-D},
    year = {2015},
    journal = {Robotics \& Automation Magazine},
    author = {Holz, Dirk and Ichim, Alexandru E and Tombari, Federico and Rusu, Radu B and Behnke, Sven},
    number = {4},
    pages = {110--124},
    volume = {22},
    publisher= {IEEE}
}

@article{gsaxner2021augmented,
  title={Augmented reality for head and neck carcinoma imaging: Description and feasibility of an instant calibration, markerless approach},
  author={Gsaxner, Christina and Pepe, Antonio and Li, Jianning and Ibrahimpasic, Una and Wallner, J{\"u}rgen and Schmalstieg, Dieter and Egger, Jan},
  journal={Computer Methods and Programs in Biomedicine},
  volume={200},
  pages={105854},
  year={2021},
  publisher={Elsevier}
}

@inproceedings{lu2019deepvcp,
  title={Deepvcp: An end-to-end deep neural network for point cloud registration},
  author={Lu, Weixin and Wan, Guowei and Zhou, Yao and Fu, Xiangyu and Yuan, Pengfei and Song, Shiyu},
  booktitle={ICCV},
  pages={12--21},
  year={2019}
}

@inproceedings{fu2021robust,
  title={Robust point cloud registration framework based on deep graph matching},
  author={Fu, Kexue and Liu, Shaolei and Luo, Xiaoyuan and Wang, Manning},
  booktitle={CVPR},
  pages={8893--8902},
  year={2021}
}

\end{document}